\documentclass{article}

\usepackage[preprint,nonatbib]{neurips_2019}

\usepackage[utf8]{inputenc} 
\usepackage[T1]{fontenc}    
\usepackage{hyperref}       
\usepackage{url}            
\usepackage{booktabs}       
\usepackage{amsfonts}       
\usepackage{nicefrac}       
\usepackage{microtype}      
\usepackage{graphicx}

\title{Prevalence of code mixing in semi-formal patient communication in low resource languages of South Africa}

\author{%
  Dr Monika Obrocka \\
  Praekelt.org\\
  Cape Town\\
  South Africa \\
  \texttt{monika@praekelt.org} \\
  \And
  Dr Charles Copley \\
  Praekelt.org\\
  Cape Town\\
  South Africa \\
  \texttt{charles@praekelt.org} \\
  \AND
  Themba Gqaza \\
  Praekelt.org\\
  Cape Town\\
  South Africa \\
  \texttt{themba@praekelt.org} \\
  \And
  Dr Eli Grant \\
  Praekelt.org\\
  Cape Town\\
  South Africa \\
  \texttt{eli@praekelt.org} \\
}

\begin{document}

\maketitle

\begin{abstract}
In this paper we address the problem of code-mixing in resource-poor language settings. We examine data consisting of 182k unique questions generated by users of the MomConnect helpdesk, part of a national scale public health platform in South Africa. We show evidence of code-switching at the level of approximately 10\% within this dataset- a level that is likely to pose challenges for future services. We use a natural language processing library (\textit{Polyglot}) that supports detection of 196 languages and attempt to evaluate its performance at identifying English, isiZulu and code-mixed questions.

\end{abstract}

\section{Introduction}
Code-mixing is a linguistic phenomenon where two languages are used spontaneously in one sentence. Code-mixing is widespread in multilingual and multicultural communities \cite{Mazibuko_2012}. South Africa is a multilingual country where the Constitution recognises 11 official languages namely Afrikaans, English, isiNdebele, isiXhosa, isiZulu, Sepedi, Sesotho, Setswana, siSwati, Tshivenda and Xitsonga with about 98\% of the total population speaking one of these as a first language \cite{census_2011}. English is often used as  a \textit{lingua franca} and dominates the published media. However, it is only the fourth most prevalent first language in the country \cite{census_2011}. Widespread use of numerous languages poses obvious challenges for the development of nationally relevant automated language processing tools. In the South African context, language tool development is further complicated by variation in each language's use across diverse socio-economic and cultural contexts. Finally, the development of human language technologies (HLT), like  corpora, lexica and software, have been hindered by significantly lower levels of digital access by those populations speaking South African languages. As a result, even the most widely spoken South African languages are classified as low-resource languages (LR). Nevertheless, it is becoming increasingly clear that the development of these tools is critically important to bridge the digital divide of a multilingual society. These tools are increasingly recognised as key  providing access to information and automated language tools. 

In this paper we highlight the challenges of an automated question-answering task for the MomConnect program run by the National Department of Health \cite{daniel-etal-2019-towards}. These are questions sent by users representing a large proportion of women attending their first antenatal care (ANC1) with registration rates increasing from  40\% in 2015, to 55\% in 2016 and to 64\% in 2017 \cite{LeFevree000583}. Amongst this population users have registered with the rates of languages~\cite{LeFevree000583} given in Table~\ref{tab:momconnect_registration_distribution}.
\begin{table}[t]
\caption{Percentage of MomConnect language registrations in 2017}
    \centering
    \begin{tabular}{l|l}
       \textbf{Language} & \textbf{Proportion} \\\hline
       English  & 55.7\% (n=745216) \\
       isiZulu & 20.3\% (n=272422) \\
       isiXhosa  & 8.4\% (n=112594) \\
       other & 15.5\% (n=207588)
    \end{tabular}
    \label{tab:momconnect_registration_distribution}
\end{table}

Linguistic analysis and  computational  modelling  is challenging alone in the LR setting, but the task is further complicated by a prevalence of code-mixing, contractions, non-standard spellings, and ungrammatical constructions in our data set. Code switching degrades the performance of natural language processing (NLP) techniques, and language identification at token level is very challenging as there are fewer features available to document level language identification.

In order to quantify the challenges, we present an analysis of the prevalence of code-switching in a data set generated on a National Health platform, MomConnect (see Table 2). The dataset is comprised of 182k unique messages with examples from each of the 11 official South African languages. We present an algorithm to identify code switching, and evaluate the performance of the algorithm by comparing it to single language identification. To our knowledge, this is the first such analysis of a national scale text programme. The South African National Health Insurance proposed for 2026 could benefit from the use of Natural Language processing and these results provide a basis for estimating the level of effort that will be required to support these services.

\begin{table}[t]
  \caption{Examples of code mixed comments.}
  \centering
  \begin{tabular}{ll}
    \toprule
    Comment     & Tag   \\
    \midrule
    \footnotesize{kuyenzeka yini kuthi umakuqhume condom kuvele kuthi khulelwe after day} & en,zu,xh \\
    \footnotesize{Mng kade ngagcina ukuthol msg evela kini}     & zu,en      \\
    \footnotesize{why ningaphenduli if umuntu ebuza something}     & zu,en         \\
    \bottomrule
  \end{tabular}
  \label{tab:code_mixing_examples}
\end{table}


\section{Methods}
\label{meth}
For this paper we focused on detecting code switching in English, isiXhosa and isiZulu, the three most common languages used in the MomConnect population \cite{LeFevree000583}. We have evaluated \textit{Polyglot} as a means of tagging languages and code-switching by comparing the automated labelling provided by \textit{Polyglot} against four manually labelled samples.  Our data was labelled by native speakers and the pre-processing consisted of the following stages:
\begin{enumerate}
    \item Removal of punctuation, emojis and digits from the data
    \item Split each question into four chunks
    \item Apply \textit{Polyglot} to each chunk
    \item Record \textit{Polyglot} label
\end{enumerate}
To evaluate the performance of this algorithm, we then compiled the following datasets.

\begin{enumerate}
    \item \textbf{Full Data Sample} 400 randomly drawn sentences ignoring the \textit{Polyglot} labels with manual language tags
    \item \textbf{English} 400 randomly drawn sentences from those tagged by \textit{Polyglot} as English with manual language tags
    \item \textbf{Zulu} 400 randomly drawn sentences from those tagged by \textit{Polyglot} as Zulu with manual language tags
    \item \textbf{Code-switched} 400 randomly drawn sentences from those tagged by \textit{Polyglot} as English~+~isiZulu, English~+~isiXhosa or isiZulu~+~isiXhosa with manual language tags
\end{enumerate}

\section{Results}
\label{analysis}

The breakdown of the different languages and language combinations is given in Table~\ref{post_lang_distr}.
\begin{table}[t]
  \caption{Manually tagged language distribution in each randomly chosen dataset of 400 samples (in \%).}
  \label{post_lang_distr}
  \centering
  \begin{tabular}{l|llll|llll}
    \toprule
    \multicolumn{1}{c}{} &  \multicolumn{4}{c}{\textbf{Multilingual}}  &  \multicolumn{4}{c}{\textbf{Monolingual}} \\
    \textbf{Dataset}     & Eng-Zu    & Eng-Xh    & Zu-Xh     & Other     & Eng   & Zu    & Xh    & Other \\
    \midrule
     Full Data Sample &4.5         &  2.75         &  0.50          &3.25           & 76.50      &4.50       &3.25       &4.50 \\
     English        & 0         & 0         & 0         & 0         & 100   & 0     & 0     & 0 \\
     isiZulu           & 11.75     & 0.4       & 7         & 0.75      & 0.4   & 75.75 & 4.25  & 0\\
     Code-switched  & 23.25     & 16        & 17.5      & 8.75      & 5.75  & 22    & 6.5   & 0.4 \\
    \bottomrule
  \end{tabular}
\end{table}
This distribution of language of incoming questions (Full Data Sample) is different to the languages chosen during registration given in Table~\ref{tab:momconnect_registration_distribution} ($\chi^{2}$=93.168, df = 3, p-value < 2.2e-16). Of the sample of 400 questions identified by the classifier as English, all were correct, whereas with isiZulu, this reduced to approximately 76\%. Code switching was present in 65.5\% of the questions that the classifier identified. An evaluation of the performance of the classifier on the  Full Data Sample gave an accuracy of 0.78, weighted precision of 0.89, and weighted recall of 0.78.

\section{Discussion}
It is interesting that the level of interaction with the service in English is higher than the level of English registrations. In addition there is evidence of extensive code switching in the data at the level of approximately 10\%.  The classifier appears to work  well in these examples as evidenced by the high rate of positive predictions evident in the \textit{English}, \textit{isiZulu}, and \textit{code-switched} data sets. However, more effort needs to be applied to evaluate the model using other techniques \cite{DBLP:journals/corr/PlataniosPMH17}. An attempt to evaluate the model when applied to the \textit{Full Data Sample} provides an accuracy of 0.775 which is not significantly higher than what would be obtained by simply assuming all questions were English (0.765). However this data set is highly imbalanced and more effort needs to be spent exploring means of evaluating the accuracy, precision and sensitivity of the model.

\section{Conclusion}
This paper demonstrates the challenges involved in natural language processing for resource poor environments at a national scale. These include imbalanced language distributions and evidence of extensive code-switching. These algorithms will need to be improved in the future to provide similar levels of digital access in these environments. A simple language classifier appears to show promise in being able to identify language and code-switching, although the evaluation of the model requires more thought given the imbalanced nature of the data set.

\subsection*{Comments}
Presented at NeurIPS 2019 Workshop on Machine Learning for the Developing World.
%

\medskip

\small

\bibliographystyle{unsrt}
\bibliography{biblio}

\end{document}